\documentclass{bmvc2k}

\usepackage{amsmath}
\usepackage{amssymb}
\usepackage{commath}
\usepackage{pifont}
\newcommand{\cmark}{\ding{51}}%
\newcommand{\xmark}{\ding{55}}%
\usepackage{array}
\newcolumntype{P}[1]{>{\centering\arraybackslash}p{#1}}
\usepackage{graphicx}
\newlength{\characterlength}
\title{Recurrent CNN for 3D Gaze Estimation using Appearance and Shape Cues}

\addauthor{Cristina Palmero\textsuperscript{1,2}}{crpalmec7@alumnes.ub.edu}{1}
\addauthor{Javier Selva\textsuperscript{1}}{javier.selva.castello@est.fib.upc.edu}{2}
\addauthor{Mohammad Ali Bagheri\textsuperscript{3,4}}{mohammadali.bagheri@ucalgary.ca}{3}
\addauthor{Sergio Escalera\textsuperscript{1,2}}{sergio@maia.ub.es}{4}

\addinstitution{
	\textsuperscript{1 }Dept. Mathematics and Informatics\\
	\hspace{\characterlength}Universitat de Barcelona, Spain
}
\addinstitution{
	\textsuperscript{2 }Computer Vision Center\\
	\hspace{\characterlength}Campus UAB, Bellaterra, Spain
}
\addinstitution{
	\textsuperscript{3 }Dept. Electrical and Computer Eng.\\
	\hspace{\characterlength}University of Calgary, Canada
}
\addinstitution{
	\textsuperscript{4 }Dept. Engineering\\
	\hspace{\characterlength}University of Larestan, Iran	
}

\runninghead{Palmero \etal}{Multi-Modal Recurrent CNN for 3D Gaze Estimation}

\def\etal{\emph{et al}\bmvaOneDot}

\begin{document}

\maketitle

\begin{abstract}
	
	Gaze behavior is an important non-verbal cue in social signal processing and human-computer interaction. In this paper, we tackle the problem of person- and head pose-independent 3D gaze estimation from remote cameras, using a multi-modal recurrent convolutional neural network (CNN). We propose to combine face, eyes region, and face landmarks as individual streams in a CNN to estimate gaze in still images. Then, we exploit the dynamic nature of gaze by feeding the learned features of all the frames in a sequence to a many-to-one recurrent module that predicts the 3D gaze vector of the last frame. Our multi-modal static solution is evaluated on a wide range of head poses and gaze directions, achieving a significant improvement of 14.6\% over the state of the art on EYEDIAP dataset, further improved by 4\% when the temporal modality is included.
	
\end{abstract}

\section{Introduction}
\label{sec:intro}
\vspace{-0.2cm}	
Eyes and their movements are considered an important cue in non-verbal behavior analysis, being involved in many cognitive processes and reflecting our internal state \cite{liversedge2000saccadic}. More specifically, eye gaze behavior, as an indicator of human visual attention, has been widely studied to assess communication skills \cite{rutter1987turn} and to identify possible behavioral disorders \cite{guillon2014visual}. Therefore, gaze estimation has become an established line of research in computer vision, being a key feature in human-computer interaction (HCI) and usability research \cite{jacob2003eye, majaranta2014eye}. 

Recent gaze estimation research has focused on facilitating its use in general everyday applications under real-world conditions, using off-the-shelf remote RGB cameras and removing the need of personal calibration \cite{palmero2018mutual}. In this setting, appearance-based methods, which learn a mapping from images to gaze directions, are the preferred choice \cite{ono2006gaze}. However, they need large amounts of training data to be able to generalize well to in-the-wild situations, which are characterized by significant variability in head poses, face appearances and lighting conditions. In recent years, CNNs have been reported to outperform classical methods. However, most existing approaches have only been tested in restricted HCI tasks, where users look at the screen or mobile phone, showing a low head pose variability. It is yet unclear how these methods would perform in a wider range of head poses. 

On a different note, until very recently, the majority of methods only used static eye region appearance as input. State-of-the-art approaches have demonstrated that using the face along with a higher resolution image of the eyes \cite{krafka2016eye}, or even just the face itself \cite{zhang2017s}, increases performance. Indeed, the whole-face image encodes more information than eyes alone, such as illumination and head pose. Nevertheless, gaze behavior is not static. Eye and head movements allow us to direct our gaze to target locations of interest. It has been demonstrated that humans can better predict gaze when being shown image sequences of other people moving their eyes \cite{anderson2016motion}. However, it is still an open question whether this sequential information can increase the performance of automatic methods. 

In this work, we show that the combination of multiple cues benefits the gaze estimation task. In particular, we use face, eye region and facial landmarks from still images. Facial landmarks model the global shape of the face and come at no cost, since face alignment is a common pre-processing step in many facial image analysis approaches. Furthermore, we present a subject-independent, free-head recurrent 3D gaze regression network to leverage the temporal information of image sequences. The static streams of each frame are combined in a late-fusion fashion using a multi-stream CNN. Then, all feature vectors are input to a many-to-one recurrent module that predicts the gaze vector of the last sequence frame. 	

In summary, our contributions are two-fold. First, we present a Recurrent-CNN network architecture that combines appearance, shape and temporal information for 3D gaze estimation. Second, we test static and temporal versions of our solution on the EYEDIAP dataset \cite{FunesMora_ETRA_2014} in a wide range of head poses and gaze directions, showing consistent performance improvements compared to related appearance-based methods. To the best of our knowledge, this is the first third-person, remote camera-based approach that uses temporal information for this task. Table \ref{tab:related} outlines our main method characteristics compared to related work. Models and code are publicly available at \url{https://github.com/crisie/RecurrentGaze}.

\section{Related work}
\label{sec:related}

\begin{table}
	\resizebox{\textwidth}{!}{%
		\begin{tabular}{|l|P{13mm}|P{17mm}|P{8mm}|P{9mm}|P{16mm}|P{17mm}|}
			\hline
			Method & 3D gaze direction & Unrestricted gaze target & Full face & Eye region & Facial landmarks & Sequential information  \\
			\hline\hline
			Zhang \etal (1) \cite{zhang2015appearance} & \cmark & \xmark & \xmark & \cmark & \xmark & \xmark \\
			Krafka \etal \cite{krafka2016eye} & \xmark & \xmark & \cmark & \cmark & \xmark & \xmark \\
			Zhang \etal (2) \cite{zhang2017s} & \cmark & \xmark & \cmark & \xmark & \xmark & \xmark \\
			Deng and Zhu \cite{deng2017monocular} & \cmark & \cmark & \cmark & \cmark & \xmark & \xmark \\
			Ours & \cmark & \cmark & \cmark & \cmark & \cmark & \cmark \\
			\hline
		\end{tabular}
	}
	\\
	\caption{Characteristics of recent related work on person- and head pose-independent appearance-based gaze estimation methods using CNNs.}
	\label{tab:related}
\end{table}

Gaze estimation methods are typically categorized as model-based or appearance-based \cite{hansen2010eye, ferhat2016low, kar2017review}. \textbf{Model-based approaches} use a geometric model of the eye, usually requiring either high resolution images or a person-specific calibration stage to estimate personal eye parameters~\cite{yoo2005novel, morimoto2002detecting, venkateswarlu2003eye, wood2014eyetab, wang2017real}. In contrast, \textbf{appearance-based methods} learn a direct mapping from intensity images or extracted eye features to gaze directions, thus being potentially applicable to relatively low resolution images and mid-distance scenarios. Different mapping functions have been explored, such as neural networks \cite{baluja1994non}, adaptive linear regression (ALR) \cite{lu2011inferring}, local interpolation \cite{tan2002appearance}, gaussian processes \cite{williams2006sparse, sugano2013appearance}, random forests \cite{huang2017tabletgaze, sugano2014learning}, or k-nearest neighbors \cite{wood2016learning}. Main challenges of appearance-based methods for 3D gaze estimation are head pose, illumination and subject invariance without user-specific calibration. To handle these issues, some works proposed compensation methods \cite{lu2011head} and warping strategies that synthesize a canonical, frontal looking view of the face \cite{mora2012gaze, funes2016gaze, jeni2016person}. Hybrid approaches based on analysis-by-synthesis have also been evaluated \cite{wood20163d}.

Currently, data-driven methods are considered the state of the art for person- and  head pose-independent appearance-based gaze estimation. Consequently, a number of gaze estimation datasets have been introduced in recent years, either in controlled \cite{smith2013gaze} or semi-controlled settings \cite{funes2014eyediap}, in the wild \cite{zhang2015appearance, krafka2016eye}, or consisting of synthetic data \cite{sugano2014learning, wood2015rendering, wood2016learning}. Zhang \etal \cite{zhang2015appearance} showed that CNNs can outperform other mapping methods, using a multi-modal CNN to learn the mapping from 3D head poses and eye images to 3D gaze directions. Krafka \etal \cite{krafka2016eye} proposed a multi-stream CNN for 2D gaze estimation, using individual eye, whole-face image and the face grid as input. As this method was limited to 2D screen mapping, Zhang \etal \cite{zhang2017s} later explored the potential of just using whole-face images as input to estimate 3D gaze directions. Using a spatial weights CNN, they demonstrated their method to be more robust to facial appearance variation caused by head pose and illumination than eye-only methods. While the method was evaluated in the wild, the subjects were only interacting with a mobile device, thus restricting the head pose range. Deng and Zhu \cite{deng2017monocular} presented a two-stream CNN to disjointly model head pose from face images and eyeball movement from eye region images. Both were then aggregated into 3D gaze direction using a gaze transform layer. The decomposition was aimed to avoid head-correlation overfitting of previous data-driven approaches. They evaluated their approach in the wild with a wider range of head poses, obtaining better performance than previous eye-based methods. However, they did not test it on public annotated benchmark datasets.

In this paper, we propose a multi-stream recurrent CNN network for person- and head pose-independent 3D gaze estimation for a mid-distance scenario. We evaluate it on a wider range of head poses and gaze directions than screen-targeted approaches. As opposed to previous methods, we also rely on temporal information inherent in sequential data. 

\section{Methodology}
\label{sec:method}

\begin{figure}
	\begin{center}
		\includegraphics[width=\columnwidth]{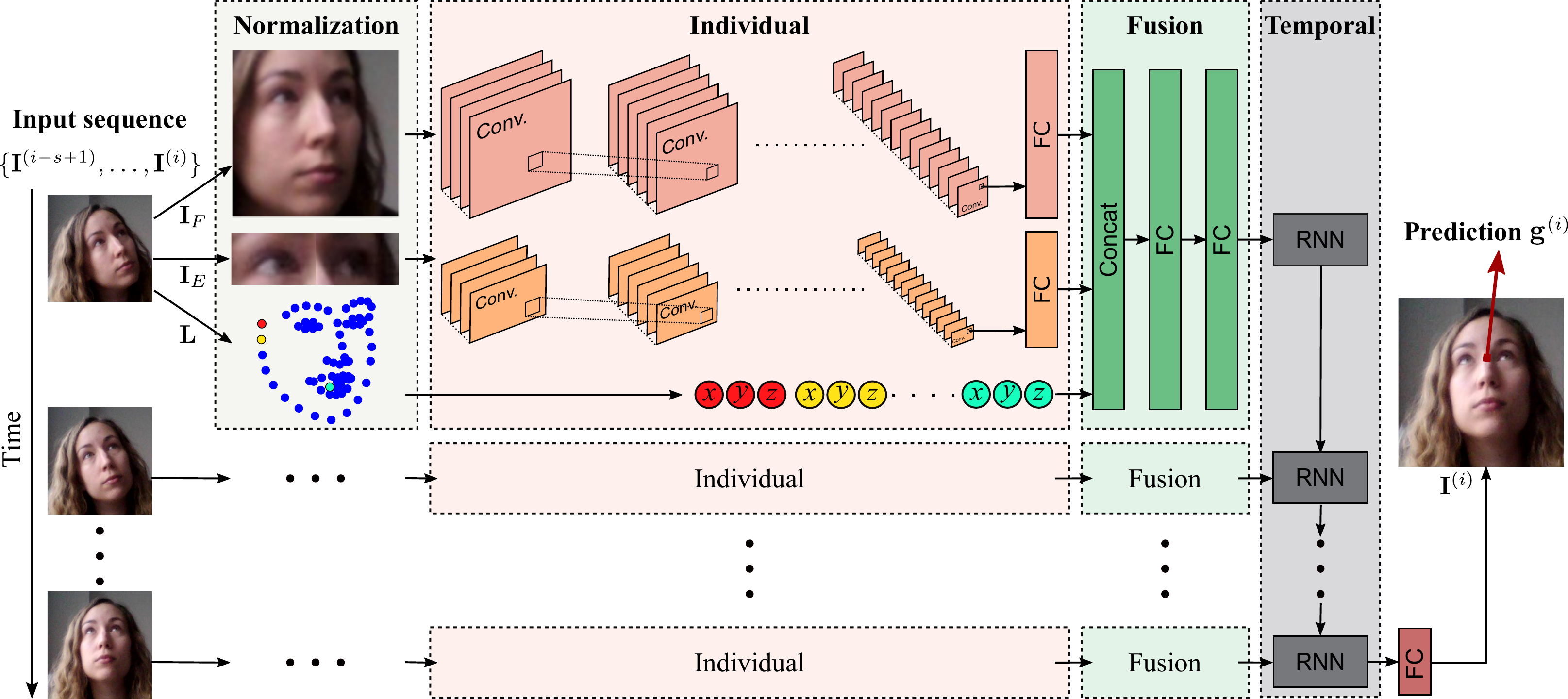}
	\end{center}
	\caption{Overview of the proposed network. A multi-stream CNN jointly models full-face, eye region appearance and face landmarks from still images. The combined extracted features from each frame are fed into a recurrent module to predict last frame's gaze direction.}
	\label{fig:architecture}
\end{figure}

In this section, we present our approach for 3D gaze regression based on appearance and shape cues for still images and image sequences. First, we introduce the data modalities and formulate the problem. Then, we detail the normalization procedure prior to the regression stage. Finally, we explain the global network topology as well as the implementation details. An overview of the system architecture is depicted in Figure \ref{fig:architecture}.

\subsection{Multi-modal gaze regression}

Let us represent gaze direction as a 3D unit vector $\mathbf{g} = [g_x, g_y, g_z]^T \in \mathbb{R}^3$ in the Camera Coordinate System (CCS), whose origin is the central point between eyeball centers. Assuming a calibrated camera, and a known head position and orientation, our goal is to estimate $\mathbf{g}$ from a sequence of images $\{\mathbf{I}^{(i)} \mid \mathbf{I} \in \mathbb{R}^{W \times H \times 3} \}$ as a regression problem. 

Gazing to a specific target is achieved by a combination of eye and head movements, which are highly coordinated. Consequently, the apparent direction of gaze is influenced not only by the location of the irises within the eyelid aperture, but also by the position and orientation of the face with respect to the camera. Known as the Wollaston effect \cite{wollaston1824xiii}, the exact same set of eyes may appear to be looking in different directions due to the surrounding facial cues. It is therefore reasonable to state that eye images are not sufficient to estimate gaze direction. Instead, whole-face images can encode head pose or illumination-specific information across larger areas than those available just in the eyes region \cite{zhang2017s, krafka2016eye}. 

The drawback of appearance-only methods is that global structure information is not explicitly considered. In that sense, facial landmarks can be used as global shape cues to encode spatial relationships and geometric constraints. Current state-of-the-art face alignment approaches are robust enough to handle large appearance variability, extreme head poses and occlusions, being especially useful when the dataset used for gaze estimation does not contain such variability. Facial landmarks are mainly correlated with head orientation, eye position, eyelid openness, and eyebrow movement, which are valuable features for our task. 

Therefore, in our approach we jointly model appearance and shape cues (see Figure \ref{fig:architecture}). The former is represented by a whole-face image $\mathbf{I}_F$, along with a higher resolution image of the eyes $\mathbf{I}_E$ to identify subtle changes. Due to dealing with wide head pose ranges, some eye images may not depict the whole eye, containing mostly background or other surrounding facial parts instead. For that reason, and contrary to previous approaches that only use one eye image \cite{zhang2015appearance, sugano2014learning}, we use a single image composed of two patches of centered left and right eyes. Finally, the shape cue is represented by 3D face landmarks obtained from a 68-landmark model, denoted by $\mathbf{L} = \{(l_x, l_y, l_z)_c \mid \forall c \in [1,...,68]\}$.

In this work we also consider the dynamic component of gaze. We leverage the sequential information of eye and head movements such that, given appearance and shape features of consecutive frames, it is possible to better predict the gaze direction of the current frame. Therefore, the 3D gaze estimation task for a 1-frame sequence is formulated as $\mathbf{g}^{(i)} = {\Large \mathnormal{f}}\left(\{{\mathbf{I}_F}^{(i)}\}, \{{\mathbf{I}_E}^{(i)}\}, \{\mathbf{L}^{(i)}\}\right)$, where $i$ denotes the $i$-th frame, and $\mathnormal{f}$ is the regression function.

\subsection{Data normalization}
\label{sec:normalization}

Prior to gaze regression, a normalization step in the 3D space and the 2D image, similar to \cite{sugano2014learning}, is carried out. This is performed to reduce the appearance variability and to allow the gaze estimation model to be applied regardless of the original camera configuration.

Let $\mathbf{H} \in \mathbb{R}^{3x3}$ be the head rotation matrix, and  $\mathbf{p} = [p_x,p_y,p_z]^T \in \mathbb{R}^3$ the reference face location with respect to the original CCS. The goal is to find the conversion matrix  $\mathbf{M} = \mathbf{S} \mathbf{R}$ such that (a) the $X$-axes of the virtual camera and the head become parallel using the rotation matrix $\mathbf{R}$, and (b) the virtual camera looks at the reference location from a fixed distance $d_n$ using the $Z$-direction scaling matrix $\mathbf{S} = diag(1,1, d_n/\lVert \mathbf{p} \rVert)$. $\mathbf{R}$ is computed as $\mathbf{a} = \mathbf{\hat{p}} \times \mathbf{H}^T \mathbf{e_1}$, $\mathbf{b} = \mathbf{\hat{a}} \times \mathbf{\hat{p}}$, $\mathbf{R} = [\mathbf{\hat{a}}, \mathbf{\hat{b}}, \mathbf{\hat{p}}]^T$, where $\mathbf{e_1}$ denotes the first orthonormal basis and $\langle \; \hat{\cdot} \; \rangle$ is the unit vector. 

This normalization translates into the image space as a cropped image patch of size $W_n \times H_n$ centered at $\mathbf{p}$ where head roll rotation has been removed. This is done by applying a perspective warping to the input image $\mathbf{I}$ using the transformation matrix $\mathbf{W} = \mathbf{C}_o\mathbf{M}{\mathbf{C}_n}^{-1}$, where $\mathbf{C}_o$ and $\mathbf{C}_n$ are the original and virtual camera matrices, respectively.

The 3D gaze vector is also normalized as $\mathbf{g}_n = \mathbf{R}\mathbf{g}$. After image normalization, the line of sight can be represented in a 2D space. Therefore, $\mathbf{g}_n$ is further transformed to spherical coordinates $(\theta,\phi)$ assuming unit length, where $\theta$ and $\phi$ denote the horizontal and vertical direction angles, respectively. This 2D angle representation, delimited in the range $[-\pi/2, \pi/2]$, is computed as $\theta = \arctan(g_x / g_z)$ and $\phi = \arcsin(-g_y)$, such that $(0,0)$ represents looking straight ahead to the CCS origin.

\subsection{Recurrent Convolutional Neural Network}
\label{sec:arch}

We propose a Recurrent CNN Regression Network for 3D gaze estimation. The network is divided in 3 modules: (1) \textit{Individual}, (2) \textit{Fusion}, and (3) \textit{Temporal}. 

First, the \textit{Individual} module learns features from each appearance cue separately. It consists of a two-stream CNN, one devoted to the normalized face image stream and the other to the joint normalized eyes image. Next, the \textit{Fusion} module combines the extracted features of each appearance stream in a single vector along with the normalized landmark coordinates. Then, it learns a joint representation between modalities in a late-fusion fashion. Both \textit{Individual} and \textit{Fusion} modules, further referred to as \textit{Static} model, are applied to each frame of the sequence. Finally, the resulting feature vectors of each frame are input to the \textit{Temporal} module based on a many-to-one recurrent network. This module leverages sequential information to predict the normalized 2D gaze angles of the last frame of the sequence using a linear regression layer added on top of it. 

\subsection{Implementation details}
\label{sec:implementation}

\subsubsection{Network details}
\label{sec:netdetails}

Each stream of the \textit{Individual} module is based on the VGG-16 deep network \cite{Parkhi15}, consisting of 13 convolutional layers, 5 max pooling layers, and 1 fully connected (FC) layer with Rectified Linear Unit (ReLU) activations. The full-face stream follows the same configuration as the base network, having an input of $224 \times 224$ pixels and a 4096D FC layer. In contrast, the input joint eye image is smaller, with a final size of $120 \times 48$ pixels, so the number of parameters is decreased proportionally. In this case, its last FC layer produces a 1536D vector. A 204D landmark coordinates vector is concatenated to the output of the FC layer of each stream, resulting in a 5836D feature vector. Consequently, the \textit{Fusion} module consists of 2 5836D FC layers with ReLU activations and 2 dropout layers between FCs as regularization. Finally, to model the temporal dependencies, we use a single GRU layer with 128 units.

The network is trained in a stage-wise fashion. First, we train the \textit{Static} model and the final regression layer end-to-end on each individual frame of the training data. The convolutional blocks are pre-trained with the VGG-Face dataset \cite{Parkhi15}, whereas the FCs are trained from scratch. Second, the training data is re-arranged by means of a sliding window with stride 1 to build input sequences. Each sequence is composed of $s = 4$ consecutive frames, whose gaze direction target is the gaze direction of the last frame of the sequence  $\left(\{\mathbf{I}^{(i-s+1)}, \dots, \mathbf{I}^{(i)} \}, \ \mathbf{g}^{(i)} \right) $. Using this re-arranged training data, we extract features of each frame of the sequence from a frozen \textit{Individual} module, fine-tune the \textit{Fusion} layers, and train both, the \textit{Temporal} module and a new final regression layer from scratch. This way, the network can exploit the temporal information to further refine the fusion weights.

We trained the model using ADAM optimizer with an initial learning rate of 0.0001, dropout of 0.3, and batch size of 64 frames. The number of epochs was experimentally set to 21 for the first training stage and 10 for the second. We use the average Euclidean distance between the predicted and ground-truth 3D gaze vectors as loss function. 

\subsubsection{Input pre-processing}
For this work we use head pose and eye locations in the 3D scene provided by the dataset. The 3D landmarks are extracted using the state-of-the-art method of Bulat and Tzimiropoulos \cite{bulat2017far}, which is based on stacked hourglass networks \cite{newell2016stacked}. 

During training, the original image is pre-processed to get the two normalized input images. The normalized whole-face patch is centered 0.1 meters ahead of the head center in the head coordinate system, and $\mathbf{C}_n$ is defined such that the image has size of $250 \times 250$ pixels. The difference between this size and the final input size allows us to perform random cropping and zooming to augment the data (explained in Section \ref{sec:dataset}). Similarly, each normalized eye patch is centered in their respective eye center locations. In this case, the virtual camera matrix is defined so that the image is cropped to $70 \times 58$, while in practice the final patches have size of $60 \times 48$. Landmarks are normalized using the same procedure and further pre-processed with mean subtraction and min-max normalization per axis. Finally, we divide them by a scaling factor $w$ such that all coordinates are in the range $[0,w]$. This way, all concatenated feature values are in a similar range. After inference, the predicted normalized 2D angles are de-normalized back to the original 3D space.

\section{Experiments}
\label{sec:exps}

\begin{figure} 
	\begin{tabular}{cccc}
		\bmvaHangBox{\includegraphics[width=3.2cm]{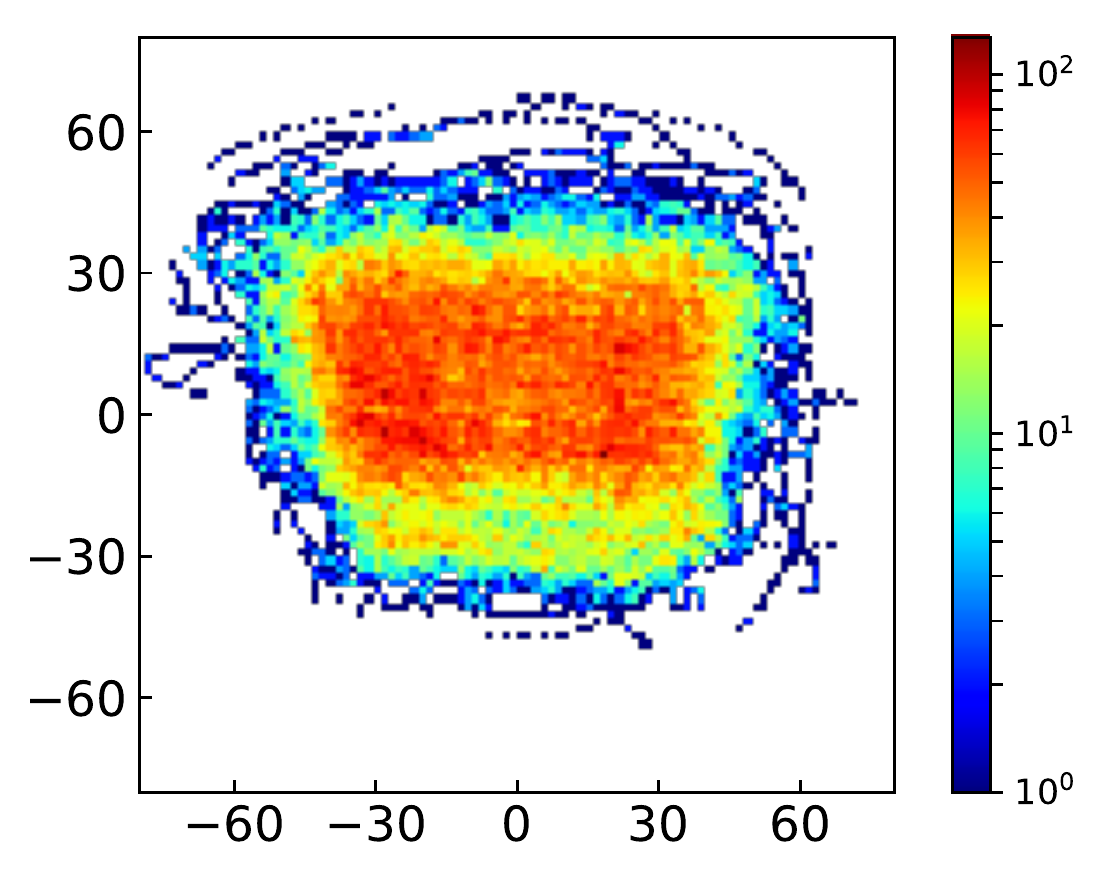}}& 
		\hspace{-0.55cm}
		\bmvaHangBox{\includegraphics[width=3.2cm]{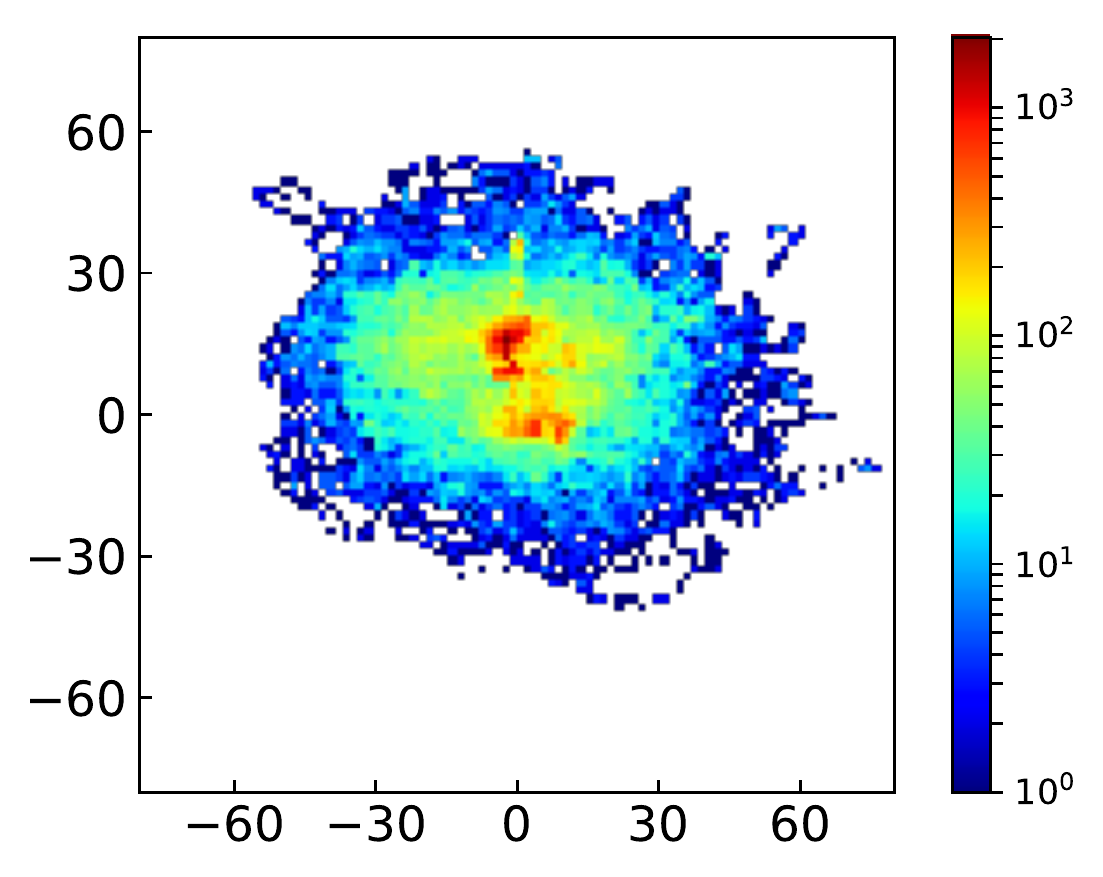}}&
		\hspace{-0.55cm}
		\bmvaHangBox{\includegraphics[width=3.2cm]{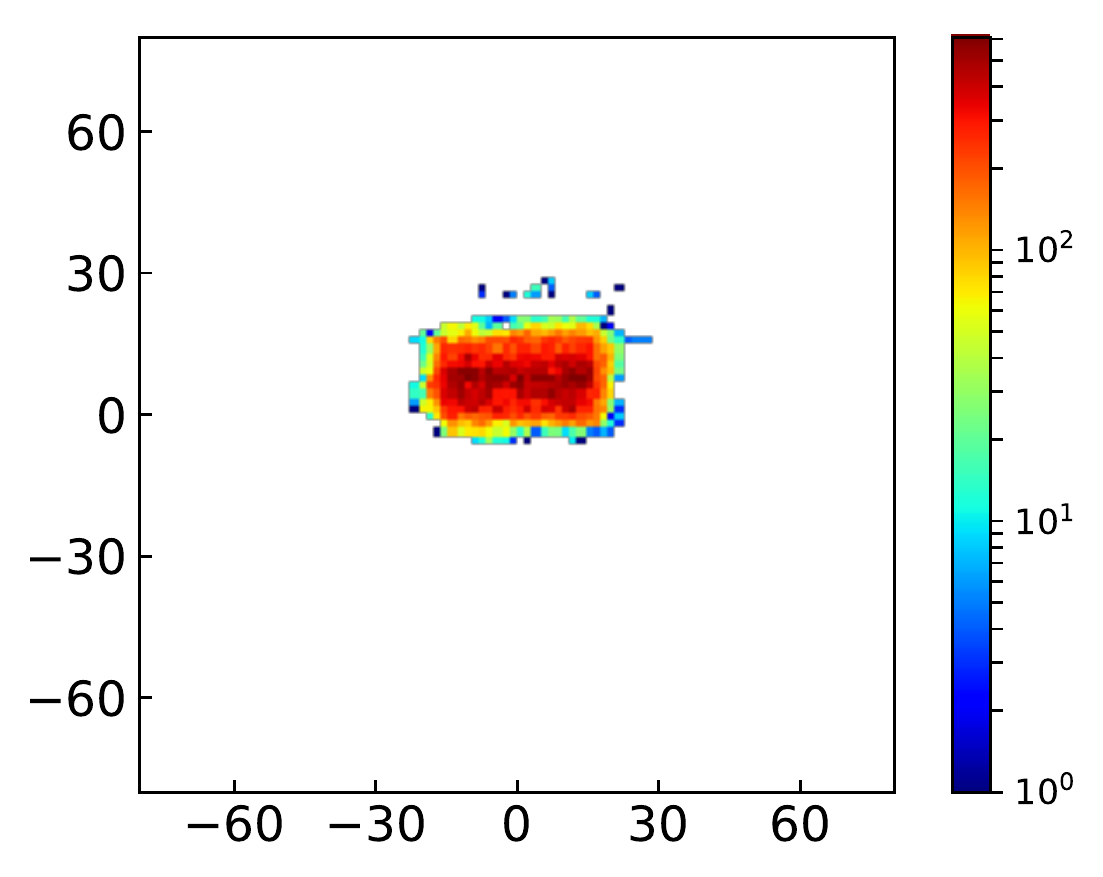}}&
		\hspace{-0.55cm}
		\bmvaHangBox{\includegraphics[width=3.2cm]{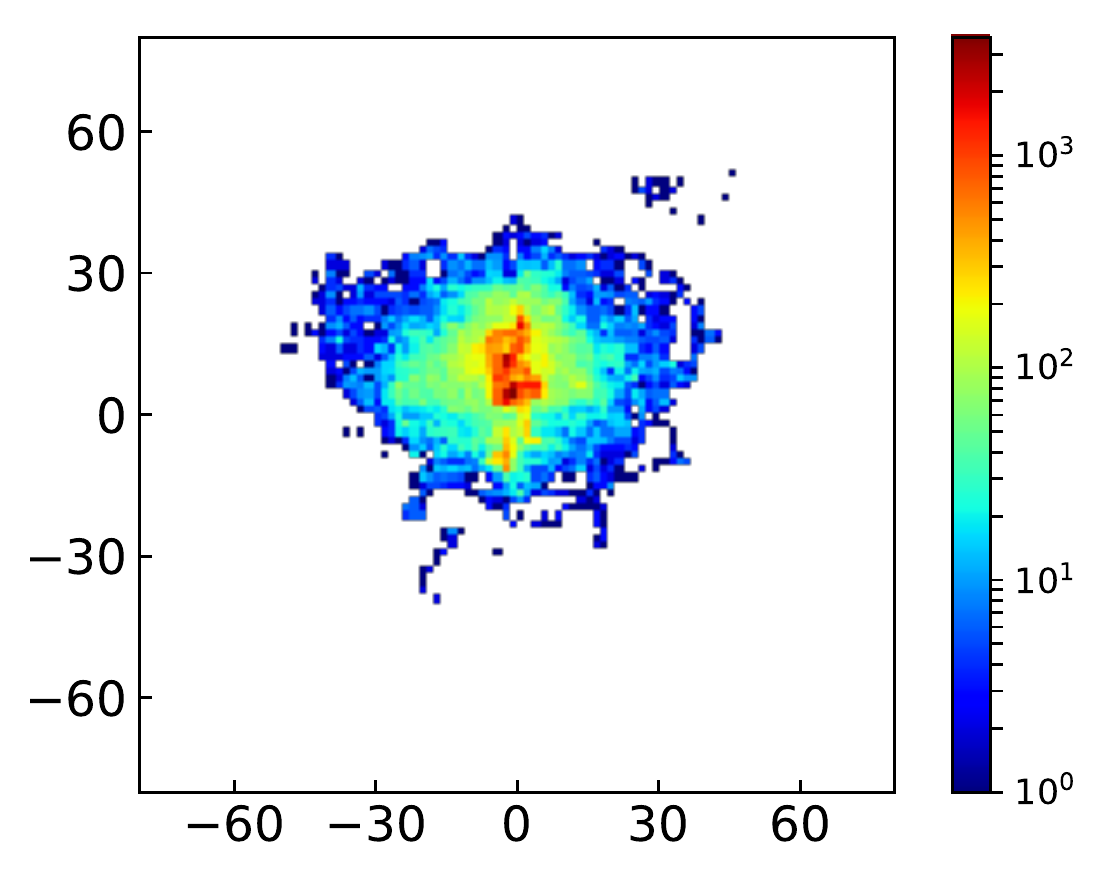}}\\
		(a) \ $\textbf{g} \ (FT)$ & \hspace{-0.55cm} (b) \ $\textbf{h} \ (FT)$ & \hspace{-0.55cm} (c) \ $\textbf{g} \ (CS)$ & \hspace{-0.55cm} (d) \ $\textbf{h} \ (CS)$ 
	\end{tabular}
	\vspace{0.01mm}
	\caption{Ground-truth eye gaze $\textbf{g}$ and head orientation $\textbf{h}$ distribution on the filtered EYEDIAP dataset for $CS$ and $FT$ settings, in terms of x- and y- angles.}
	\label{fig:eyediap_dist}
\end{figure}

In this section, we evaluate the cross-subject 3D gaze estimation task on a wide range of head poses and gaze directions. Furthermore, we validate the effectiveness of the proposed architecture comparing both static and temporal approaches. We report the error in terms of mean angular error between predicted and ground-truth 3D gaze vectors. Note that due to the requirements of the temporal model not all the frames obtain a prediction. Therefore, for a fair comparison, the reported results for static models disregard such frames when temporal models are included in the comparison.

\subsection{Training data}
\label{sec:dataset}

There are few publicly available datasets devoted to 3D gaze estimation and most of them focus on HCI with a limited range of head pose and gaze directions. Therefore, we use VGA videos from the publicly-available EYEDIAP dataset \cite{FunesMora_ETRA_2014} to perform the experimental evaluation, as it is currently the only one containing video sequences with a wide range of head poses and showing the full face. This dataset consists of 3-minute videos of 16 subjects looking at two types of targets: continuous \textit{screen} targets on a fixed monitor ($CS$), and \textit{floating} physical targets ($FT$). The videos are further divided into \textit{static} ($S$) and \textit{moving} ($M$) head pose for each of the subjects. Subjects 12-16 were recorded with 2 different lighting conditions.

For evaluation, we filtered out those frames that fulfilled at least one of the following conditions: (1) face or landmarks not detected; (2) subject not looking at the target; (3) 3D head pose, eyes or target location not properly recovered; and (4) eyeball rotations violating physical constraints ($\abs{\theta} \le 40^\circ$, $\abs{\phi} \le 30^\circ$) \cite{msccirc}. Note that we purposely do not filter eye blinking moments to learn their dynamics with the temporal model, which may produce some outliers with a higher prediction error due to a less accurate ground truth. Figure \ref{fig:eyediap_dist} shows the distribution of gaze directions and head poses for both filtered $CS$ and $FT$ cases.

We applied data augmentation to the training set with the following random transformations:  horizontal flip, shifts of up to 5 pixels, zoom of up to 2\%, brightness changes by a factor in the range $[0.4, 1.75$], and additive Gaussian noise with $\sigma^2 = 0.03$.

\subsection{Evaluation of static modalities}

First, we evaluate the contribution of each static modality on the $FT$ scenario. We divided the 16 participants into 4 groups, such that appearance variability was maximized while maintaining a similar number of training samples per group. Each static model was trained end-to-end performing 4-fold cross-validation using different combinations of input modalities. Since the number of fusion units depends on the number of input modalities, we also compare different fusion layer sizes. The effect of data normalization is also evaluated by training a not-normalized face model where the input image is the face bounding box with square size the maximum distance between 2D landmarks. 

\begin{figure}
	\begin{minipage}{.48\columnwidth}
		\centering
		\bmvaHangBox{\includegraphics[width=5.5cm]{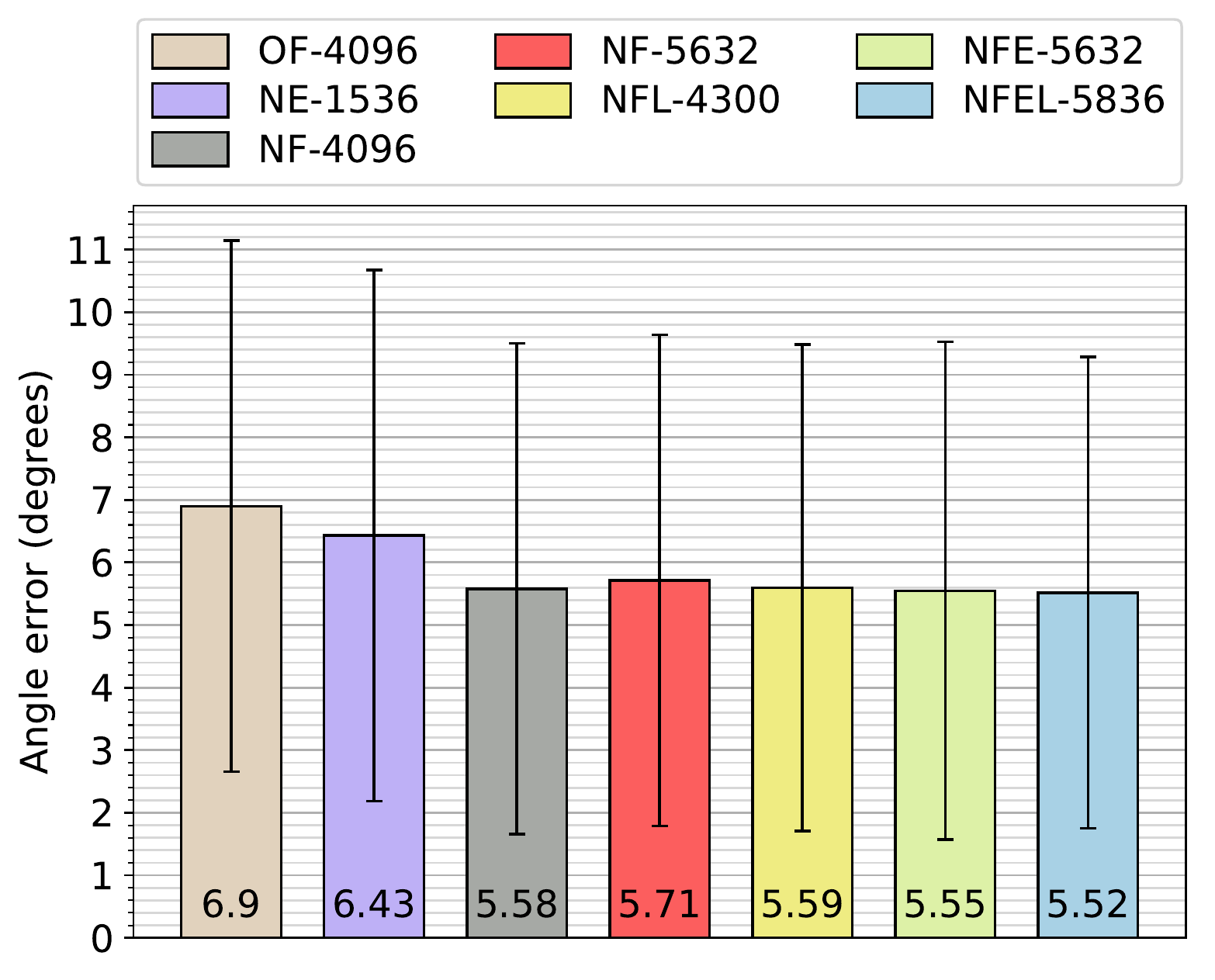}}
		\caption{Performance evaluation of the \textit{Static} network using different input modalities (\textit{O - Not normalized, N - Normalized, F - Face, E - Eyes, L - 3D Landmarks}) and size of fusion layers on the \textit{FT} scenario.}
		\label{fig:static_comparison}
	\end{minipage}
	\hspace{.03\columnwidth}
	\begin{minipage}{.48\columnwidth}
		\centering
		\bmvaHangBox{\includegraphics[width=5.5cm]{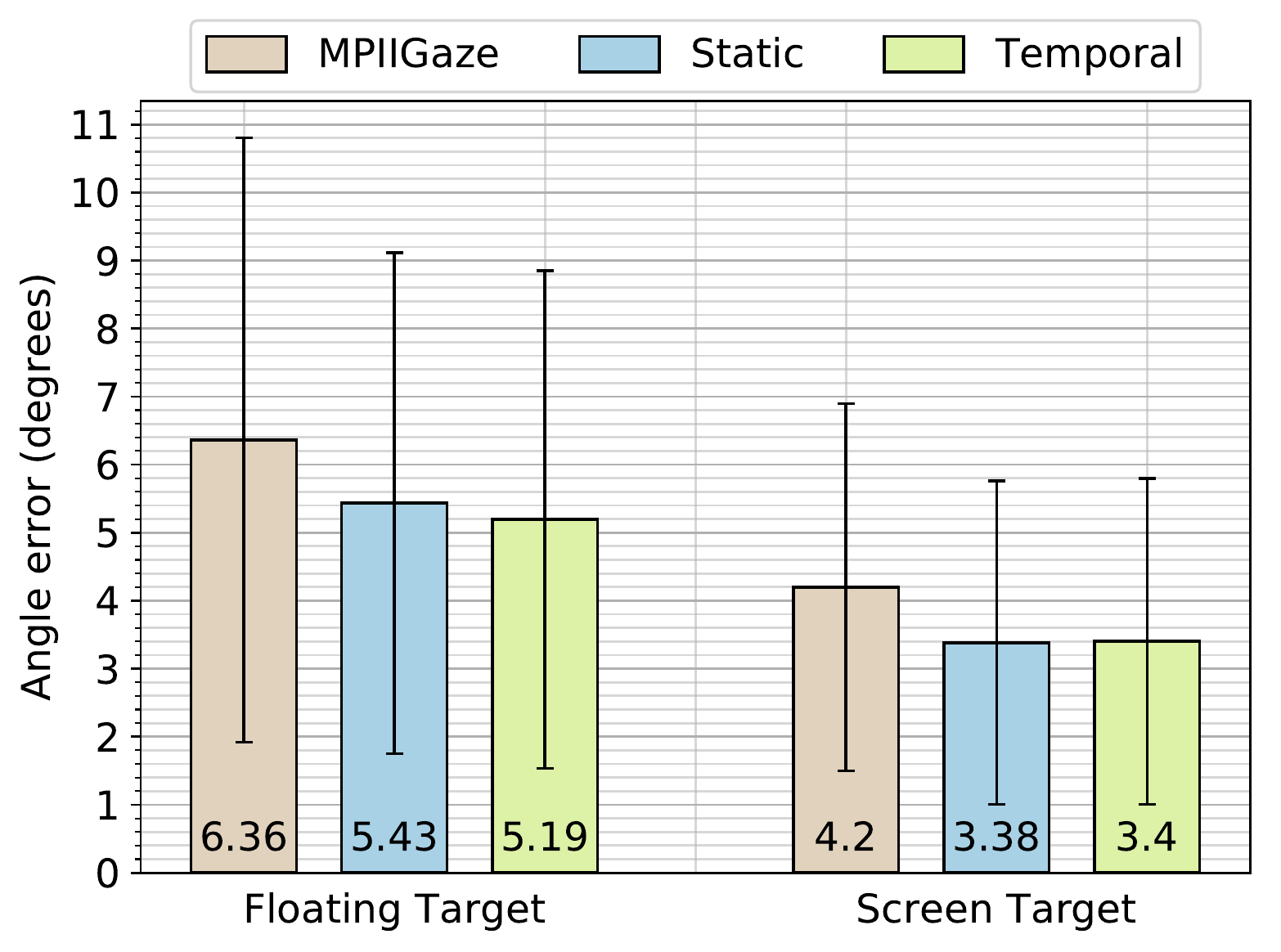}}
		\caption{Performance comparison among MPIIGaze method \cite{zhang2015appearance} and our \textit{Static} and \textit{Temporal} versions of the proposed network for $FT$ and $CS$ scenarios.}
		\label{fig:sota_comparison}
	\end{minipage}
\end{figure}

As shown in Figure \ref{fig:static_comparison}, all models that take normalized full-face information as input achieve better performance than the eyes-only model. More specifically, the combination of face, eyes and landmarks outperforms all the other combinations by a small but significant margin (paired Wilcoxon test, $p < 0.0001$). The standard deviation of the best-performing model is reduced compared to the face and eyes model, suggesting a regularizing effect due to the addition of landmarks. The not-normalized face-only model shows the largest error, proving the impact of normalization to reduce the appearance variability. Furthermore, our results indicate that the increase of fusion units is not correlated with a better performance.

\subsection{Static gaze regression: comparison with existing methods}
\label{sec:comparison_sota}

We compare our best-performing static model with three baselines. \textbf{Head:} Treating the head pose directly as gaze direction. \textbf{PR-ALR:} Method that relies on RGB-D data to rectify the eye images viewpoint into a canonical head pose using a 3DMM. It then learns an RGB gaze appearance model using ALR \cite{mora2012gaze}. Predicted 3D vectors for \textit{FT-S} scenario are provided by EYEDIAP dataset. \textbf{MPIIGaze:}. State-of-the-art full-face 3D gaze estimation method \cite{zhang2015appearance}. They use an Alexnet-based CNN model with spatial weights to enhance information in different facial regions. We fine-tuned it with the filtered EYEDIAP subsets using our training parameters and normalization procedure. 

In addition to the aforementioned \textit{FT}-based evaluation setup, we also evaluate our method on the \textit{CS} scenario. In this case there are only 14 participants available, so we divided them in 5 groups and performed 5-fold cross-validation. In Figure \ref{fig:sota_comparison} we compare our method to MPIIGaze, achieving a statistically significant improvement of 14.6\% and 19.5\% on \textit{FT} and \textit{CS} scenarios, respectively (paired Wilcoxon test, $p < 0.0001$). We can observe that a restricted gaze target benefits the performance of all methods, compared to a more challenging unrestricted setting with a wider range of head poses and gaze directions.

Table \ref{tab:leaveoneout} provides a detailed comparison on every participant, performing leave-one-out cross-validation on the \textit{FT} scenario for \textit{static} and \textit{moving} head separately. Results show that, as expected, facial appearance and head pose have a noticeable impact on gaze accuracy, with average error differences of up to $7.7^\circ$ among participants.  

\begin{table}
	\resizebox{\textwidth}{!}{%
		\begin{tabular}{|l|c|c|c|c|c|c|c|c|c|c|c|c|c|c|c|c|c|}
			\hline
			Method & 1 & 2 & 3 & 4 & 5 & 6 & 7 & 8 & 9 & 10 & 11 & 12 & 13 & 14 & 15 & 16 & Avg. \\
			\hline\hline
			Head & 23.5 & 22.1 & 20.3 & 23.6 & 23.2 & 23.2 & 23.6 & 21.2 & 26.7 & 23.6 & 23.1 & 24.4 & 23.3 & 24.0 & 24.5 & 22.8 & 23.3 \\
			PR-ALR & 12.3 & 12.0 & 12.4 & 11.3 & 15.5 & 12.9 & 17.9 & 11.8 & 17.3 & 13.4 & 13.4 & 14.3 & 15.2 & 13.6 & 14.4 & 14.6 & 13.9 \\
			MPIIGaze & 5.3 & 5.1 & 5.7 & 4.7 & 7.3 & 15.1 & 10.8 & 5.7 & 9.9 & 7.1 & 5.0 & 5.7 & 7.4 & 3.8 & \textbf{4.8} & 5.5 & 6.8 \\
			Static & \textbf{3.9} & \textbf{4.1} & \textbf{4.2} & \textbf{3.9} & \textbf{6.0} & \textbf{6.4} & 7.2 & \textbf{3.6} & \textbf{7.1} & \textbf{5.0} & 5.7 &  6.7 & \textbf{3.9} & 4.7 & 5.1 & \textbf{4.2} & \textbf{5.1} \\
			Temporal & 4.0 & 4.9 & 4.3 & 4.1 & 6.1 & 6.5 & \textbf{6.6} & 3.9 & 7.8 & 6.1 & \textbf{4.7} & \textbf{5.6} & 4.7 & \textbf{3.5} & 5.9 & 4.6 & 5.2 \\
			\hline\hline
			Head & 19.3 & 14.2 & 16.4 & 19.9 & 16.8 & 21.9 & 16.1 & 24.2 & 20.3 & 19.9 & 18.8 & 22.3 & 18.1 & 14.9 & 16.2 & 19.3 & 18.7 \\
			MPIIGaze & 7.6 & 6.2 & 5.7 & 8.7 & 10.1 & 12.0 & 12.2 & 6.1 & 8.3 & 5.9 & 6.1 & 6.2 & 7.4 & 4.7 & 4.4 & 6.0 & 7.3 \\
			Static & \textbf{5.8} & 5.7 & \textbf{4.4} & \textbf{7.5} & 6.7 & 8.8 & \textbf{11.6} & 5.5 & 8.3 & 5.5 & 5.2 & 6.3 & \textbf{5.3} & \textbf{3.9} & \textbf{4.3} & \textbf{5.6} & 6.3 \\
			Temporal & 6.1 & \textbf{5.6} & 4.5 & \textbf{7.5} & \textbf{6.4} & \textbf{8.2} & 12.0 & \textbf{5.0} & \textbf{7.5} & \textbf{5.4} & \textbf{5.0} & \textbf{5.8} & 6.6 & 4.0 & 4.5 & 5.8 & \textbf{6.2} \\
			\hline
		\end{tabular}
	}
	\\
	\caption{Gaze angular error comparison for \textit{static} (top half) and \textit{moving} (bottom half) head pose for each subject in the $FT$ scenario. Best results in bold.}
	\label{tab:leaveoneout}
\end{table}

\subsection{Evaluation of the temporal network}
\label{sec:temporal}

\begin{figure}
	\begin{tabular}{cccc}
		\bmvaHangBox{\includegraphics[width=3.15cm]{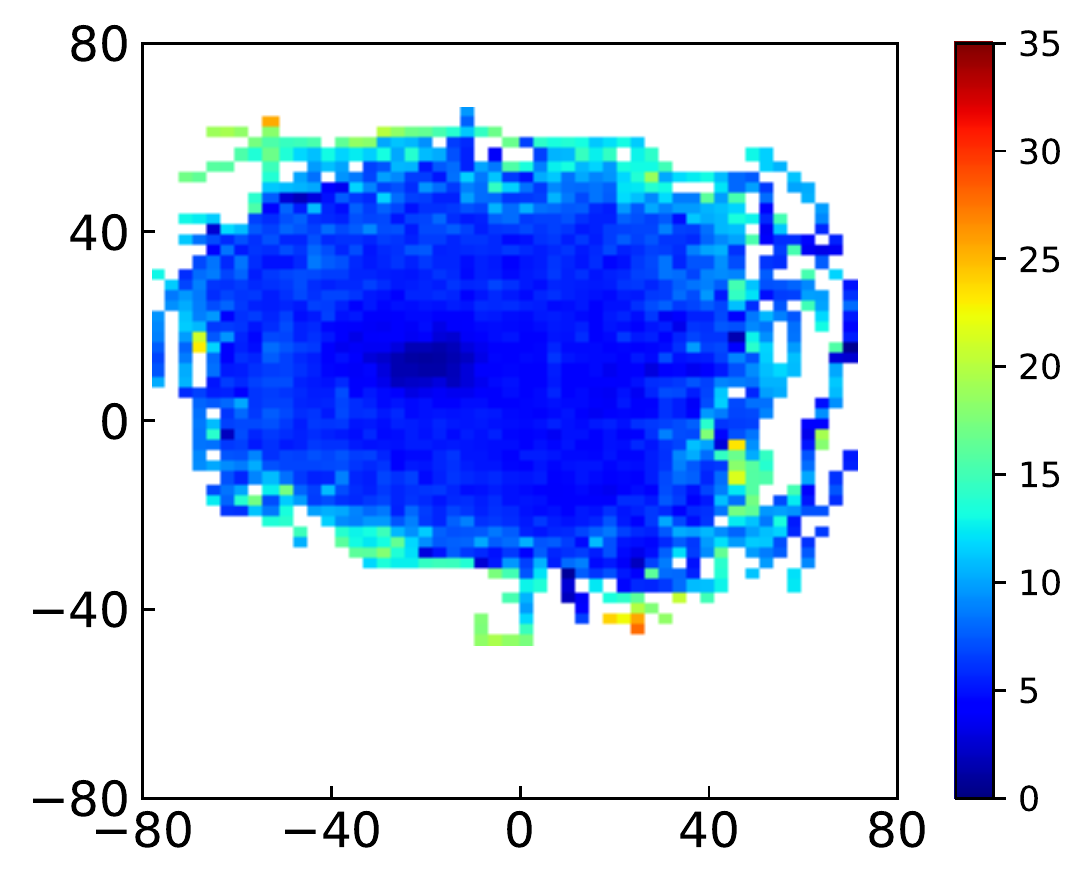}}&
		\hspace{-0.5cm}
		\bmvaHangBox{\includegraphics[width=3.15cm]{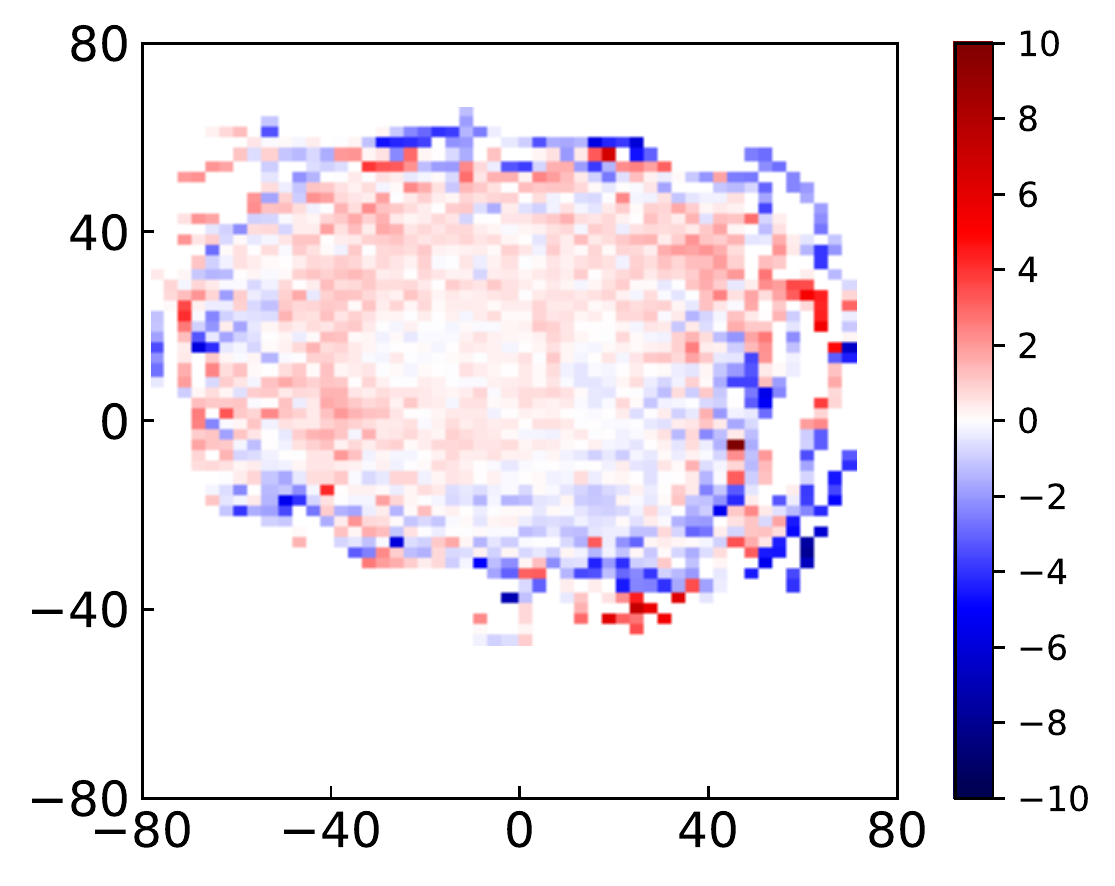}}&
		\hspace{-0.4cm}
		\bmvaHangBox{\includegraphics[width=3.15cm]{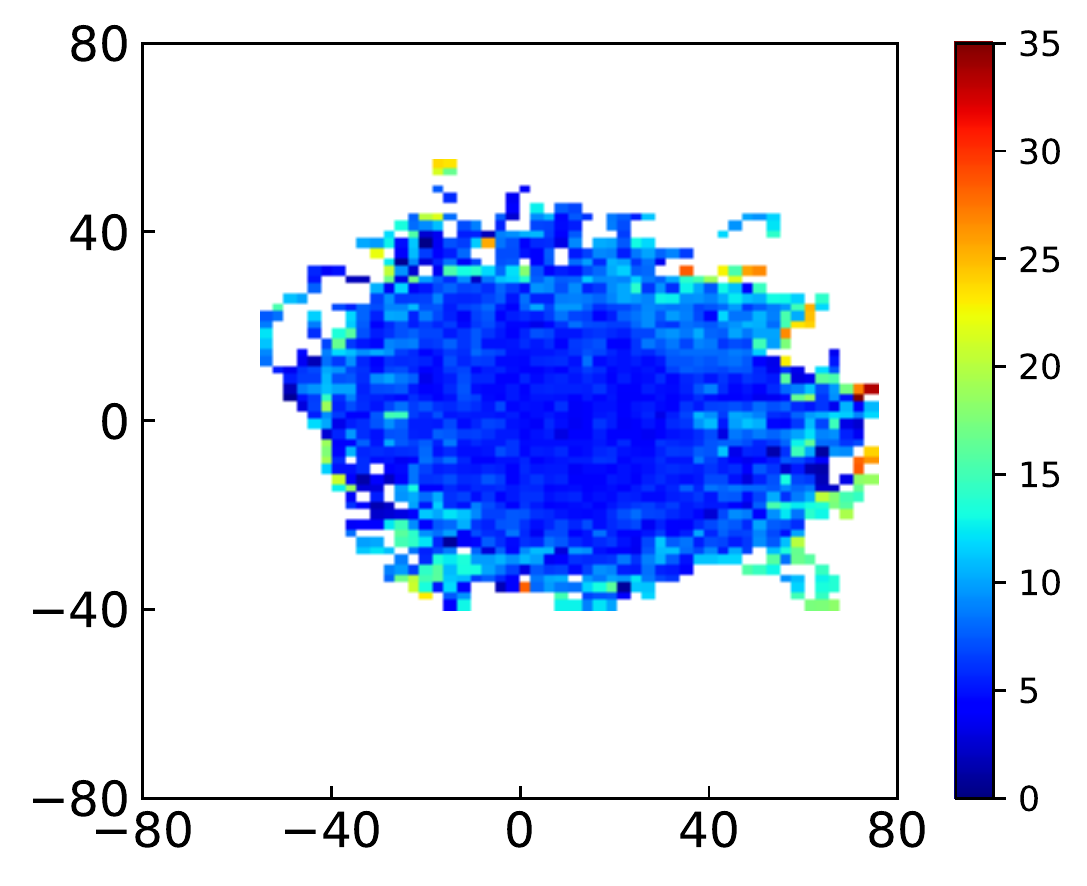}}&
		\hspace{-0.5cm}
		\bmvaHangBox{\includegraphics[width=3.15cm]{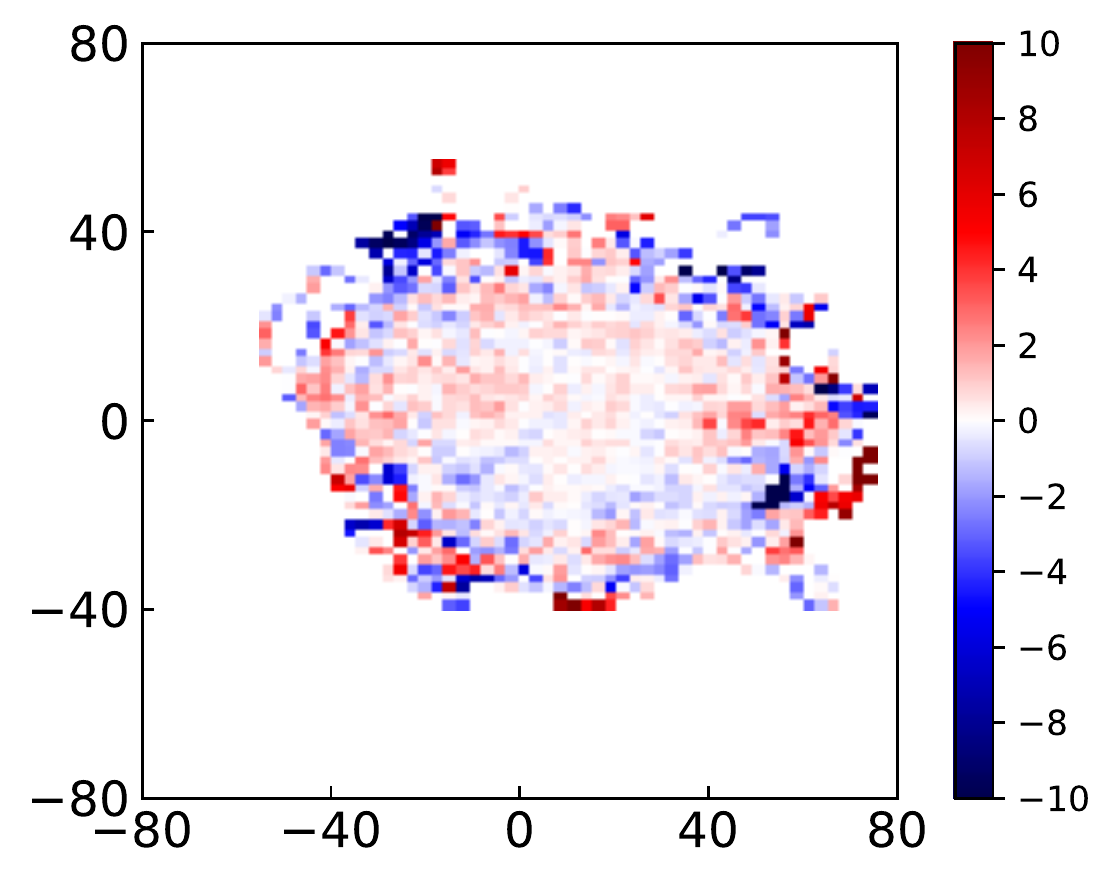}}\\
	\end{tabular}
	\begin{tabular}{cc}
		\hspace{2.cm} (a) Gaze space & \hspace{2.8cm} (b) Head orientation space\\
	\end{tabular}
	\caption{Angular error distribution across gaze (a) and head orientation (b) spaces in the \textit{FT} setting, in terms of x- and y- angles. For each space, we depict the \textit{Static} model performance (left) and the contribution of the \textit{Temporal} model versus \textit{Static} (right). In the latter, positive difference means higher improvement of the \textit{Temporal} model.}
	\label{fig:staticvstemporal}
\end{figure}

In this section, we evaluate the contribution of adding the temporal module to the static model. To do so, we trained a lower-dimensional version of the static network with comparable performance to the original, reducing the number of units of the second fusion layer to 2918. Results are reported in Figure \ref{fig:sota_comparison} and Table \ref{tab:leaveoneout}. One can observe that using sequential information is helpful on the \textit{FT} scenario, outperforming the static model by a statistically significant 4.4\% (paired Wilcoxon test, $p < 0.0001$). This contribution is more noticeable in the \textit{moving} head setting, proving that the temporal model can benefit from head motion information. In contrast, such information seems to be less meaningful in the \textit{CS} scenario, where the obtained error is already very low for a cross-subject setting and the amount of head movement declines.

Figure \ref{fig:staticvstemporal} further explores the error distribution of the static network and the impact of sequential information. We can observe that the accuracy of the static model drops with extreme head poses and gaze directions, which can also be correlated to having less data in those areas. Compared to the static model, the temporal model particularly benefits gaze targets from mid-range upwards. Its contribution is less clear for extreme targets, probably again due to data imbalance. 

Finally, we evaluated the effect of different recurrent architectures for the temporal model. In particular, we tested 1 (128 units) and 2 (256-128 units) LSTM and GRU layers, with 1 GRU layer obtaining slightly superior results (up to $0.12^\circ$). We also assessed the effect of sequence length fixing $s$ in the range $\{4,7,10\}$, with $s = 7$ performing worse than the other two (up to $0.14^\circ$). 

\section{Conclusions}
\label{sec:conclusions}
	
In this work, we studied the combination of full-face and eye images along with facial landmarks for person- and head pose-independent 3D gaze estimation. Consequently, we proposed a multi-stream recurrent CNN network that leverages the sequential information of eye and head movements. Both static and temporal versions of our approach significantly outperform current state-of-the-art 3D gaze estimation methods on a wide range of head poses and gaze directions. We showed that adding geometry features to appearance-based methods has a regularizing effect on the accuracy. Adding sequential information further benefits the final performance compared to static-only input, especially from mid-range upwards and in those cases where head motion is present. The effect in very extreme head poses is not clear due to data imbalance, suggesting the importance of learning from a continuous, balanced dataset including all head poses and gaze directions of interest. To the best of our knowledge, this is the first attempt to exploit the temporal modality in the context of gaze estimation from remote cameras. As future work, we will further explore extracting meaningful temporal representations of gaze dynamics, considering 3DCNNs as well as the encoding of deep features around particular tracked face landmarks \cite{jung2015joint}.

\section*{Acknowledgements}
This work has been partially supported by the Spanish project TIN2016-74946-P (MINECO/ FEDER, UE), CERCA Programme / Generalitat de Catalunya, and the FP7 people program (Marie Curie Actions), REA grant agreement no FP7-607139 (iCARE - Improving Children Auditory REhabilitation). We gratefully acknowledge the support of NVIDIA Corporation with the donation of the GPU used for this research. Portions of the research in this paper used the EYEDIAP dataset made available by the Idiap Research Institute, Martigny, Switzerland.

\bibliography{egbib}

\begin{thebibliography}{43}
\providecommand{\natexlab}[1]{#1}
\providecommand{\url}[1]{\texttt{#1}}
\expandafter\ifx\csname urlstyle\endcsname\relax
  \providecommand{\doi}[1]{doi: #1}\else
  \providecommand{\doi}{doi: \begingroup \urlstyle{rm}\Url}\fi

\bibitem[Anderson et~al.(2016)Anderson, Risko, and
  Kingstone]{anderson2016motion}
Nicola~C Anderson, Evan~F Risko, and Alan Kingstone.
\newblock Motion influences gaze direction discrimination and disambiguates
  contradictory luminance cues.
\newblock \emph{Psychonomic bulletin \& review}, 23\penalty0 (3):\penalty0
  817--823, 2016.

\bibitem[Baluja and Pomerleau(1994)]{baluja1994non}
Shumeet Baluja and Dean Pomerleau.
\newblock Non-intrusive gaze tracking using artificial neural networks.
\newblock In \emph{Advances in Neural Information Processing Systems}, pages
  753--760, 1994.

\bibitem[Bulat and Tzimiropoulos(2017)]{bulat2017far}
Adrian Bulat and Georgios Tzimiropoulos.
\newblock How far are we from solving the 2d \& 3d face alignment problem? (and
  a dataset of 230,000 3d facial landmarks).
\newblock In \emph{International Conference on Computer Vision}, 2017.

\bibitem[Deng and Zhu(2017)]{deng2017monocular}
Haoping Deng and Wangjiang Zhu.
\newblock Monocular free-head 3d gaze tracking with deep learning and geometry
  constraints.
\newblock In \emph{Computer Vision (ICCV), 2017 IEEE International Conference
  on}, pages 3162--3171. IEEE, 2017.

\bibitem[Ferhat and Vilari{\~n}o(2016)]{ferhat2016low}
Onur Ferhat and Fernando Vilari{\~n}o.
\newblock Low cost eye tracking.
\newblock \emph{Computational intelligence and neuroscience}, 2016:\penalty0
  17, 2016.

\bibitem[Funes-Mora and Odobez(2016)]{funes2016gaze}
Kenneth~A Funes-Mora and Jean-Marc Odobez.
\newblock Gaze estimation in the {3D} space using {RGB-D} sensors.
\newblock \emph{International Journal of Computer Vision}, 118\penalty0
  (2):\penalty0 194--216, 2016.

\bibitem[Funes~Mora et~al.(2014{\natexlab{a}})Funes~Mora, Monay, and
  Odobez]{FunesMora_ETRA_2014}
Kenneth~Alberto Funes~Mora, Florent Monay, and Jean-Marc Odobez.
\newblock Eyediap: A database for the development and evaluation of gaze
  estimation algorithms from rgb and rgb-d cameras.
\newblock In \emph{Proceedings of the ACM Symposium on Eye Tracking Research
  and Applications}. ACM, March 2014{\natexlab{a}}.
\newblock \doi{10.1145/2578153.2578190}.

\bibitem[Funes~Mora et~al.(2014{\natexlab{b}})Funes~Mora, Monay, and
  Odobez]{funes2014eyediap}
Kenneth~Alberto Funes~Mora, Florent Monay, and Jean-Marc Odobez.
\newblock Eyediap: A database for the development and evaluation of gaze
  estimation algorithms from rgb and rgb-d cameras.
\newblock In \emph{Proceedings of the Symposium on Eye Tracking Research and
  Applications}, pages 255--258. ACM, 2014{\natexlab{b}}.

\bibitem[Guillon et~al.(2014)Guillon, Hadjikhani, Baduel, and
  Rog{\'e}]{guillon2014visual}
Quentin Guillon, Nouchine Hadjikhani, Sophie Baduel, and Bernadette Rog{\'e}.
\newblock Visual social attention in autism spectrum disorder: Insights from
  eye tracking studies.
\newblock \emph{Neuroscience \& Biobehavioral Reviews}, 42:\penalty0 279--297,
  2014.

\bibitem[Hansen and Ji(2010)]{hansen2010eye}
Dan~Witzner Hansen and Qiang Ji.
\newblock In the eye of the beholder: A survey of models for eyes and gaze.
\newblock \emph{IEEE transactions on pattern analysis and machine
  intelligence}, 32\penalty0 (3):\penalty0 478--500, 2010.

\bibitem[Huang et~al.(2017)Huang, Veeraraghavan, and
  Sabharwal]{huang2017tabletgaze}
Qiong Huang, Ashok Veeraraghavan, and Ashutosh Sabharwal.
\newblock Tabletgaze: dataset and analysis for unconstrained appearance-based
  gaze estimation in mobile tablets.
\newblock \emph{Machine Vision and Applications}, 28\penalty0 (5-6):\penalty0
  445--461, 2017.

\bibitem[Jacob and Karn(2003)]{jacob2003eye}
Robert~JK Jacob and Keith~S Karn.
\newblock Eye tracking in human-computer interaction and usability research:
  Ready to deliver the promises.
\newblock In \emph{The mind's eye}, pages 573--605. Elsevier, 2003.

\bibitem[Jeni and Cohn(2016)]{jeni2016person}
L{\'a}szl{\'o}~A Jeni and Jeffrey~F Cohn.
\newblock Person-independent 3d gaze estimation using face frontalization.
\newblock In \emph{Proceedings of the IEEE Conference on Computer Vision and
  Pattern Recognition Workshops}, pages 87--95, 2016.

\bibitem[Jung et~al.(2015)Jung, Lee, Yim, Park, and Kim]{jung2015joint}
Heechul Jung, Sihaeng Lee, Junho Yim, Sunjeong Park, and Junmo Kim.
\newblock Joint fine-tuning in deep neural networks for facial expression
  recognition.
\newblock In \emph{Computer Vision (ICCV), 2015 IEEE International Conference
  on}, pages 2983--2991. IEEE, 2015.

\bibitem[Kar and Corcoran(2017)]{kar2017review}
Anuradha Kar and Peter Corcoran.
\newblock A review and analysis of eye-gaze estimation systems, algorithms and
  performance evaluation methods in consumer platforms.
\newblock \emph{IEEE Access}, 5:\penalty0 16495--16519, 2017.

\bibitem[Krafka et~al.(2016)Krafka, Khosla, Kellnhofer, Kannan, Bhandarkar,
  Matusik, and Torralba]{krafka2016eye}
Kyle Krafka, Aditya Khosla, Petr Kellnhofer, Harini Kannan, Suchendra
  Bhandarkar, Wojciech Matusik, and Antonio Torralba.
\newblock Eye tracking for everyone.
\newblock In \emph{Computer Vision and Pattern Recognition (CVPR), 2016 IEEE
  Conference on}, pages 2176--2184. IEEE, 2016.

\bibitem[Liversedge and Findlay(2000)]{liversedge2000saccadic}
Simon~P Liversedge and John~M Findlay.
\newblock Saccadic eye movements and cognition.
\newblock \emph{Trends in cognitive sciences}, 4\penalty0 (1):\penalty0 6--14,
  2000.

\bibitem[Lu et~al.(2011{\natexlab{a}})Lu, Okabe, Sugano, and Sato]{lu2011head}
Feng Lu, Takahiro Okabe, Yusuke Sugano, and Yoichi Sato.
\newblock A head pose-free approach for appearance-based gaze estimation.
\newblock In \emph{BMVC}, pages 1--11, 2011{\natexlab{a}}.

\bibitem[Lu et~al.(2011{\natexlab{b}})Lu, Sugano, Okabe, and
  Sato]{lu2011inferring}
Feng Lu, Yusuke Sugano, Takahiro Okabe, and Yoichi Sato.
\newblock Inferring human gaze from appearance via adaptive linear regression.
\newblock In \emph{Computer Vision (ICCV), 2011 IEEE International Conference
  on}, pages 153--160. IEEE, 2011{\natexlab{b}}.

\bibitem[Majaranta and Bulling(2014)]{majaranta2014eye}
P{\"a}ivi Majaranta and Andreas Bulling.
\newblock Eye tracking and eye-based human--computer interaction.
\newblock In \emph{Advances in physiological computing}, pages 39--65.
  Springer, 2014.

\bibitem[Mora and Odobez(2012)]{mora2012gaze}
Kenneth Alberto~Funes Mora and Jean-Marc Odobez.
\newblock Gaze estimation from multimodal kinect data.
\newblock In \emph{Computer Vision and Pattern Recognition Workshops (CVPRW),
  2012 IEEE Computer Society Conference on}, pages 25--30. IEEE, 2012.

\bibitem[Morimoto et~al.(2002)Morimoto, Amir, and
  Flickner]{morimoto2002detecting}
Carlos~Hitoshi Morimoto, Arnon Amir, and Myron Flickner.
\newblock Detecting eye position and gaze from a single camera and 2 light
  sources.
\newblock In \emph{Pattern Recognition, 2002. Proceedings. 16th International
  Conference on}, volume~4, pages 314--317. IEEE, 2002.

\bibitem[MSC()]{msccirc}
IMO MSC.
\newblock Circ. 982 (2000) guidelines on ergonomic criteria for bridge
  equipment and layout.

\bibitem[Newell et~al.(2016)Newell, Yang, and Deng]{newell2016stacked}
Alejandro Newell, Kaiyu Yang, and Jia Deng.
\newblock Stacked hourglass networks for human pose estimation.
\newblock In \emph{European Conference on Computer Vision}, pages 483--499.
  Springer, 2016.

\bibitem[Ono et~al.(2006)Ono, Okabe, and Sato]{ono2006gaze}
Yasuhiro Ono, Takahiro Okabe, and Yoichi Sato.
\newblock Gaze estimation from low resolution images.
\newblock In \emph{Pacific-Rim Symposium on Image and Video Technology}, pages
  178--188. Springer, 2006.

\bibitem[Palmero et~al.(2018)Palmero, van Dam, Escalera, Kelia, Lichtert,
  Noldus, Spink, and van Wieringen]{palmero2018mutual}
Cristina Palmero, Elisabeth~A. van Dam, Sergio Escalera, Mike Kelia, Guido~F.
  Lichtert, Lucas~P.J.J Noldus, Andrew~J. Spink, and Astrid van Wieringen.
\newblock Automatic mutual gaze detection in face-to-face dyadic interaction
  videos.
\newblock In \emph{Proceedings of Measuring Behavior}, pages 158--163, 2018.

\bibitem[Parkhi et~al.(2015)Parkhi, Vedaldi, and Zisserman]{Parkhi15}
Omkar~M. Parkhi, Andrea Vedaldi, and Andrew Zisserman.
\newblock Deep face recognition.
\newblock In \emph{British Machine Vision Conference}, 2015.

\bibitem[Rutter and Durkin(1987)]{rutter1987turn}
Derek~R Rutter and Kevin Durkin.
\newblock Turn-taking in mother--infant interaction: An examination of
  vocalizations and gaze.
\newblock \emph{Developmental psychology}, 23\penalty0 (1):\penalty0 54, 1987.

\bibitem[Smith et~al.(2013)Smith, Yin, Feiner, and Nayar]{smith2013gaze}
Brian~A Smith, Qi~Yin, Steven~K Feiner, and Shree~K Nayar.
\newblock Gaze locking: passive eye contact detection for human-object
  interaction.
\newblock In \emph{Proceedings of the 26th annual ACM symposium on User
  interface software and technology}, pages 271--280. ACM, 2013.

\bibitem[Sugano et~al.(2013)Sugano, Matsushita, and Sato]{sugano2013appearance}
Yusuke Sugano, Yasuyuki Matsushita, and Yoichi Sato.
\newblock Appearance-based gaze estimation using visual saliency.
\newblock \emph{IEEE transactions on pattern analysis and machine
  intelligence}, 35\penalty0 (2):\penalty0 329--341, 2013.

\bibitem[Sugano et~al.(2014)Sugano, Matsushita, and Sato]{sugano2014learning}
Yusuke Sugano, Yasuyuki Matsushita, and Yoichi Sato.
\newblock Learning-by-synthesis for appearance-based 3d gaze estimation.
\newblock In \emph{Computer Vision and Pattern Recognition (CVPR), 2014 IEEE
  Conference on}, pages 1821--1828. IEEE, 2014.

\bibitem[Tan et~al.(2002)Tan, Kriegman, and Ahuja]{tan2002appearance}
Kar-Han Tan, David~J Kriegman, and Narendra Ahuja.
\newblock Appearance-based eye gaze estimation.
\newblock In \emph{Applications of Computer Vision, 2002.(WACV 2002).
  Proceedings. Sixth IEEE Workshop on}, pages 191--195. IEEE, 2002.

\bibitem[Venkateswarlu et~al.(2003)]{venkateswarlu2003eye}
Ronda Venkateswarlu et~al.
\newblock Eye gaze estimation from a single image of one eye.
\newblock In \emph{Computer Vision, 2003. Proceedings. Ninth IEEE International
  Conference on}, pages 136--143. IEEE, 2003.

\bibitem[Wang and Ji(2017)]{wang2017real}
Kang Wang and Qiang Ji.
\newblock Real time eye gaze tracking with 3d deformable eye-face model.
\newblock In \emph{Proceedings of the IEEE Conference on Computer Vision and
  Pattern Recognition}, pages 1003--1011, 2017.

\bibitem[Williams et~al.(2006)Williams, Blake, and Cipolla]{williams2006sparse}
Oliver Williams, Andrew Blake, and Roberto Cipolla.
\newblock Sparse and semi-supervised visual mapping with the s\^{} 3gp.
\newblock In \emph{Computer Vision and Pattern Recognition, 2006 IEEE Computer
  Society Conference on}, volume~1, pages 230--237. IEEE, 2006.

\bibitem[Wollaston et~al.(1824)]{wollaston1824xiii}
William~Hyde Wollaston et~al.
\newblock Xiii. on the apparent direction of eyes in a portrait.
\newblock \emph{Philosophical Transactions of the Royal Society of London},
  114:\penalty0 247--256, 1824.

\bibitem[Wood and Bulling(2014)]{wood2014eyetab}
Erroll Wood and Andreas Bulling.
\newblock Eyetab: Model-based gaze estimation on unmodified tablet computers.
\newblock In \emph{Proceedings of the Symposium on Eye Tracking Research and
  Applications}, pages 207--210. ACM, 2014.

\bibitem[Wood et~al.(2015)Wood, Baltrusaitis, Zhang, Sugano, Robinson, and
  Bulling]{wood2015rendering}
Erroll Wood, Tadas Baltrusaitis, Xucong Zhang, Yusuke Sugano, Peter Robinson,
  and Andreas Bulling.
\newblock Rendering of eyes for eye-shape registration and gaze estimation.
\newblock In \emph{Proceedings of the IEEE International Conference on Computer
  Vision}, pages 3756--3764, 2015.

\bibitem[Wood et~al.(2016{\natexlab{a}})Wood, Baltru{\v{s}}aitis, Morency,
  Robinson, and Bulling]{wood20163d}
Erroll Wood, Tadas Baltru{\v{s}}aitis, Louis-Philippe Morency, Peter Robinson,
  and Andreas Bulling.
\newblock A 3d morphable eye region model for gaze estimation.
\newblock In \emph{European Conference on Computer Vision}, pages 297--313.
  Springer, 2016{\natexlab{a}}.

\bibitem[Wood et~al.(2016{\natexlab{b}})Wood, Baltru{\v{s}}aitis, Morency,
  Robinson, and Bulling]{wood2016learning}
Erroll Wood, Tadas Baltru{\v{s}}aitis, Louis-Philippe Morency, Peter Robinson,
  and Andreas Bulling.
\newblock Learning an appearance-based gaze estimator from one million
  synthesised images.
\newblock In \emph{Proceedings of the Ninth Biennial ACM Symposium on Eye
  Tracking Research \& Applications}, pages 131--138. ACM, 2016{\natexlab{b}}.

\bibitem[Yoo and Chung(2005)]{yoo2005novel}
Dong~Hyun Yoo and Myung~Jin Chung.
\newblock A novel non-intrusive eye gaze estimation using cross-ratio under
  large head motion.
\newblock \emph{Computer Vision and Image Understanding}, 98\penalty0
  (1):\penalty0 25--51, 2005.

\bibitem[Zhang et~al.(2015)Zhang, Sugano, Fritz, and
  Bulling]{zhang2015appearance}
Xucong Zhang, Yusuke Sugano, Mario Fritz, and Andreas Bulling.
\newblock Appearance-based gaze estimation in the wild.
\newblock In \emph{Proceedings of the IEEE Conference on Computer Vision and
  Pattern Recognition}, pages 4511--4520, 2015.

\bibitem[Zhang et~al.(2017)Zhang, Sugano, Fritz, and Bulling]{zhang2017s}
Xucong Zhang, Yusuke Sugano, Mario Fritz, and Andreas Bulling.
\newblock It's written all over your face: Full-face appearance-based gaze
  estimation.
\newblock In \emph{Proc. IEEE International Conference on Computer Vision and
  Pattern Recognition Workshops (CVPRW)}, 2017.

\end{thebibliography}
\end{document}